\documentclass[letterpaper, 10 pt, conference]{ieeeconf}  

\IEEEoverridecommandlockouts                              

\usepackage{graphicx} 
\graphicspath{{./pdf/}} 
\DeclareGraphicsExtensions{.pdf,jpg}
\usepackage[export]{adjustbox} 
\usepackage[skip=1pt,labelformat=empty]{subcaption} 
\usepackage[cmex10]{amsmath} 
\usepackage[short]{optidef} 
\usepackage[ruled,linesnumbered]{algorithm2e} 
\usepackage{caption} 
\usepackage{bm} 
\usepackage{epsfig} 
\usepackage{times} 
\usepackage{color} 
\usepackage{amssymb}  
\usepackage{setspace}

\newcommand{\mytilde}{\raise.17ex\hbox{$\scriptstyle\mathtt{‌​\sim}$}}
\usepackage{xparse,soul}
\usepackage{multirow}
\usepackage{cite}
\usepackage[font=small,labelfont=bf]{caption}

\usepackage[utf8]{inputenc}
\usepackage{mathtools}

\DeclareMathOperator*{\argmin}{arg\,min}

\title{\LARGE \bf
Autonomous Intelligent Navigation for Flexible Endoscopy Using Monocular Depth Guidance and 3-D Shape Planning
}

\author{Yiang Lu$^{1}$\textsuperscript{\textdagger}, Ruofeng Wei$^{2}$\textsuperscript{\textdagger}, Bin Li$^{1}$, Wei Chen$^{1}$, Jianshu Zhou$^{1,4}$, Qi Dou$^{3,4}$, Dong Sun$^{2}$, and Yun-hui Liu$^{1,4}$
\thanks{This work is supported in part by Shenzhen Portion of Shenzhen-Hong Kong Science and Technology Innovation Cooperation Zone under HZQB-KCZYB-20200089, in part of the HK RGC under T42-409/18-R and 14202918, in part by the Hong Kong Centre for Logistics Robotics, in part by the Multi-Scale Medical Robotics Centre, InnoHK, and in part by the VC Fund 4930745 of the CUHK T Stone Robotics Institute. \textit{(Corresponding author: Jianshu Zhou and Yun-Hui Liu.)}}%
\thanks{$^{1}$T Stone Robotics Institute, Department of Mechanical and Automation Engineering, The Chinese University of Hong Kong, Hong Kong.}
\thanks{$^{2}$Department of Biomedical Engineering, City University of Hong Kong, Hong Kong.}
\thanks{$^{3}$Department of Computer Science and Engineering, The Chinese University of Hong Kong, Hong Kong.}
\thanks{$^{4}$Hong Kong Center for Logistics Robotics, Hong Kong.}
\thanks{\textsuperscript{\textdagger}Y. Lu and R. Wei  contributed equally to this work.}
}

\begin{document}

\maketitle
\thispagestyle{empty}
\pagestyle{empty}

\begin{abstract}
Recent advancements toward perception and decision-making of flexible endoscopes have shown great potential in computer-aided surgical interventions.
However, owing to modeling uncertainty and inter-patient anatomical variation in flexible endoscopy, the challenge remains for efficient and safe navigation in patient-specific scenarios.
This paper presents a novel data-driven framework with self-contained visual-shape fusion for autonomous intelligent navigation of flexible endoscopes requiring no priori knowledge of system models and global environments.
A learning-based adaptive visual servoing controller is proposed to online update the eye-in-hand vision-motor configuration and steer the endoscope, which is guided by monocular depth estimation via a vision transformer (ViT).
To prevent unnecessary and excessive interactions with surrounding anatomy, an energy-motivated shape planning algorithm is introduced through entire endoscope 3-D proprioception from embedded fiber Bragg grating (FBG) sensors. 
Furthermore, a model predictive control (MPC) strategy is developed to minimize the elastic potential energy flow and simultaneously optimize the steering policy.
Dedicated navigation experiments on a robotic-assisted flexible endoscope with an FBG fiber in several phantom environments demonstrate the effectiveness and adaptability of the proposed framework.
\end{abstract}

\section{INTRODUCTION}
Robotic endoscope technologies have been widely deployed in a variety of diagnoses and treatments to ease medical devices' accessibility and alleviate physicians' burden.
Before the future prevalence of wireless endoscopy using capsule devices or magnetic manipulation \cite{martin2020enabling,prendergast2020real}, traditional flexible endoscopes, such as gastroscopes, colonoscopes, and bronchoscopes, are utilized in dominant endoscopic interventions for their cost-effectiveness and reliability \cite{van2013towards}.
Currently, endoscopic navigation based on offline trajectory and searching algorithms is well-studied  \cite{huang2021autonomous,favaro2021evolutionary}. However, flexible endoscope operation on patient-specific and unstructured scenarios is still challenging considering unknown priori environmental information, nonlinear system characteristics, and decision-making safety issues \cite{prendergast2020real,girerd2020slam}.

\begin{figure*}[!t]
  \centering
  \includegraphics[width=0.95\linewidth]{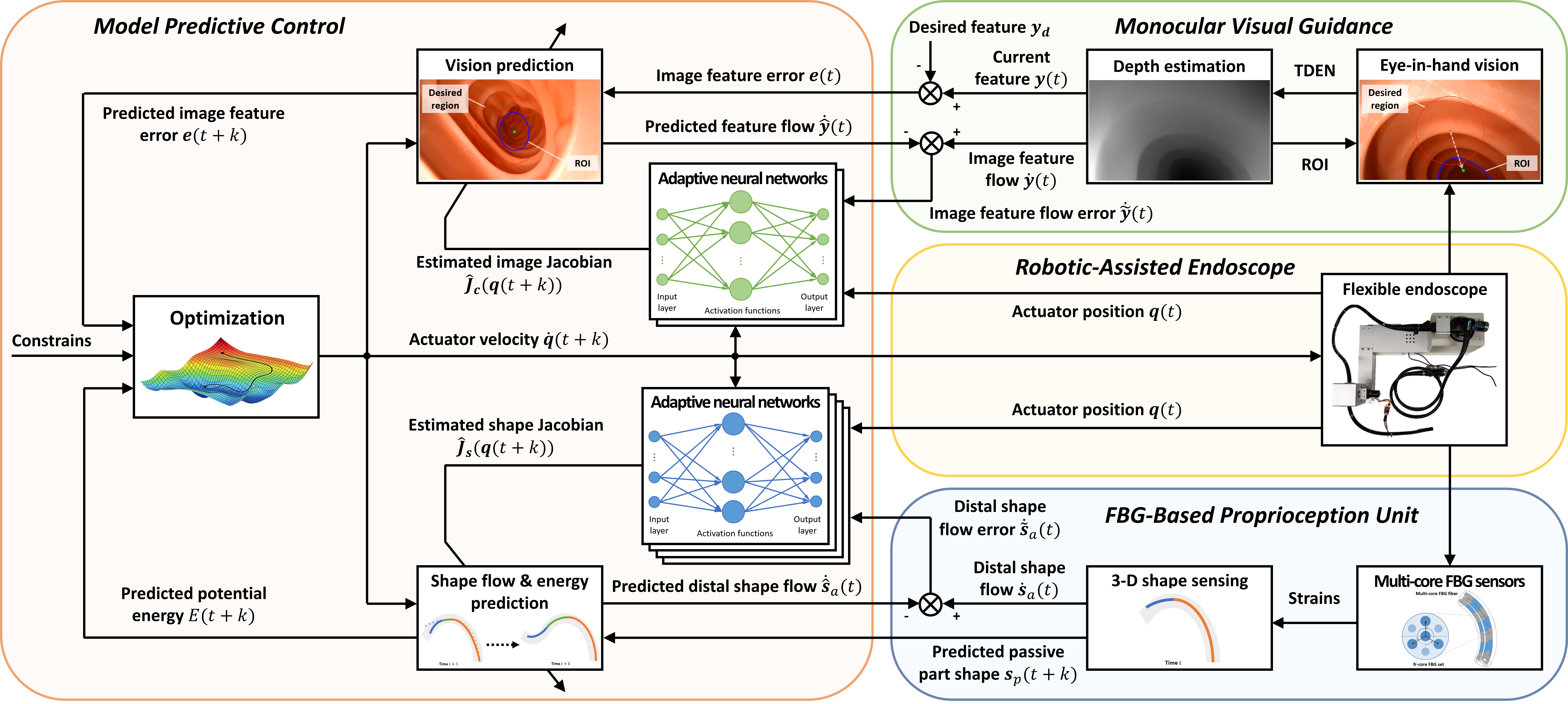}
  \setlength{\abovecaptionskip}{0.1cm}
  \caption{The proposed data-driven framework for autonomous navigation of flexible endoscopes, including robotic-assisted endoscope, monocular visual guidance, FBG-based proprioception unit, and learning-based model predictive control.}
  \label{fig:diagram}
  \vspace{-0.4cm}
\end{figure*}

Usually, endoscopic navigation leverages close visualization of intracorporeal scenes from the embedded camera, and utilizes lumen centralization and feature tracking for primary guidance \cite{van2013towards,girerd2020slam,pore2022colonoscopy,wei2022stereo}.
The lumen center and visual target can be determined by dark region segmentation \cite{martin2020enabling,pore2022colonoscopy}, depth estimation \cite{wei2022stereo}, or contours detection \cite{prendergast2020real}.
In real applications, dark region segmentation is impaired by complex lighting conditions, while versatile occlusions limit contours detection effectiveness \cite{prendergast2020real}.
Depth estimation methods with higher reliability can be mainly classified into multi-view stereo methods, learning-based approaches, and structured light solutions \cite{maurice2012structured,leonard2018evaluation,liu2019dense,girerd2020slam,li2021data,ozyoruk2021endoslam,wei2022stereo,wei2022distilled}.
Multi-view stereo methods can reconstruct 3-D scenes with depth estimation while requiring distinguishable features \cite{leonard2018evaluation,girerd2020slam}.
Learning-based approaches demonstrate their empowered intelligence, among which self-supervised and unsupervised algorithms have been investigated for depth estimation of endoscopic scenes \cite{liu2019dense,ozyoruk2021endoslam,wei2022stereo}.
However, these methods either cannot compute the depth map in real time or adopt binocular images for estimation when perceiving the 3-D scene structure.
Recently, vision transformer (ViT) shows great potential in image processing tasks, which extracts features without explicit downsampling and has a global receptive field through all stages \cite{ranftl2021vision}, thus benefiting depth estimation.

After acquiring the effective visual target as guidance and feedback, various planning and control methodologies have been developed for visual servoing of flexible endoscopes \cite{nazari2022visual}.
For model-based methods, nonlinear structural properties and unknown disturbances can result in unknown deviations in the priori system modeling. 
Model-less and learning-based strategies show great potentials for this task, which update the robot behavior with eye-in-hand vision configuration by estimation and learning approaches \cite{george2018control,fang2019vision,wang2021survey}.
Amongst, neural networks (NNs) were commonly utilized to learn the system models for visual servoing of soft and continuum manipulators \cite{wang2020eye,fang2021soft}.
Deep reinforcement learning was also investigated for image-based control of colonoscopy navigation by devising an end-to-end policy \cite{pore2022colonoscopy}.
Although they alleviate priori modeling, data exploration and offline training are required \cite{wang2021survey}.  
In addition to vision-based control, follow-the-leader deployment maintaining the endoscope shape while maneuvering the tip, is desired for flexible endoscopic navigation \cite{girerd2020slam,culmone2021follow}.

To further reduce the patient's discomfort and prevent damage to anatomical tissue resulting from unexpected interactions, force sensing and haptic guidance were deployed for monitoring the cognitive load during autonomous navigation of flexible endoscopy \cite{huang2021autonomous,reilink2011evaluation}.
Alternatively, deflection and external force in confined spaces can be estimated based on the principle of the minimum total potential energy \cite{sarli2017minimal,kim2022sigmoidal}.
Fiber Bragg grating (FBG) sensors have been widely utilized for tip localization and shape sensing of continuum medical robots \cite{wang2016shape,shi2016shape,alambeigi2019scade,wang2020eye,lu2021robust,sefati2021dexterous,lu2022fbg,cao2022end,lu2023robust}, because of small size, high biocompatibility, and high sampling rate without line-of-sight constrains.
With the working principle of strain measurements, FBGs can be also used to acquire the elastic potential energy of flexible endoscopes from distributed bending/torsion signals \cite{wang2016shape,lu2021robust}.
However, studies about energy prediction and 3-D shape planning to avoid excessive contact forces on surrounding anatomy during flexible endoscope navigation have not been reported.

In this paper, we propose a novel data-driven framework as shown in Fig. \ref{fig:diagram}, which can automatically navigate flexible endoscopes without priori data exploration and offline trajectory.
To the best of our knowledge, this is the first work in consideration of minimizing potential energy flow for planning and autonomous navigation of unmodeled endoscopes, by incorporating learning-based monocular visual guidance and control, together with FBG-based proprioception of 3-D configuration. 
The proposed method can effectively online learn the vision-motor behavior with the convergences of image tracking error and learning parameters, and adaptively compensate for the modeling uncertainty with disturbances.
Our main contributions are summarized as follows:
\begin{enumerate}
    \item Design and implementation of a ViT-based depth estimation network for monocular endoscopic guidance, and a learning-based adaptive visual servoing strategy to online update the eye-in-hand vision-motor configuration of the flexible endoscope, and simultaneously track the image region of interest (ROI).
    \item Development of an energy-motivated shape planning algorithm by leveraging FBG-based proprioception, which can not only measure the endoscope 3-D configuration but also monitor and minimize its potential energy flow to improve the navigation performance.
    \item Integration of depth guidance with learning-based shape flow and energy prediction into a model predictive control (MPC) framework for steering policy optimization of autonomous endoscope navigation.
    \item Experimental validations on a robotic-assisted colonoscope system embedded with FBG sensors in several phantoms, the results of which demonstrate the feasibility and adaptability of the proposed framework. 
\end{enumerate}

\section{Monocular Depth-Guided Visual Servoing}
This section introduces the depth-guided visual servoing strategy, including ViT-based depth estimation and image-based data-driven control of flexible endoscopes.

\subsection{ViT-Based Monocular Depth Guidance}
\begin{figure}[!t]
  \centering
  \includegraphics[width=\linewidth]{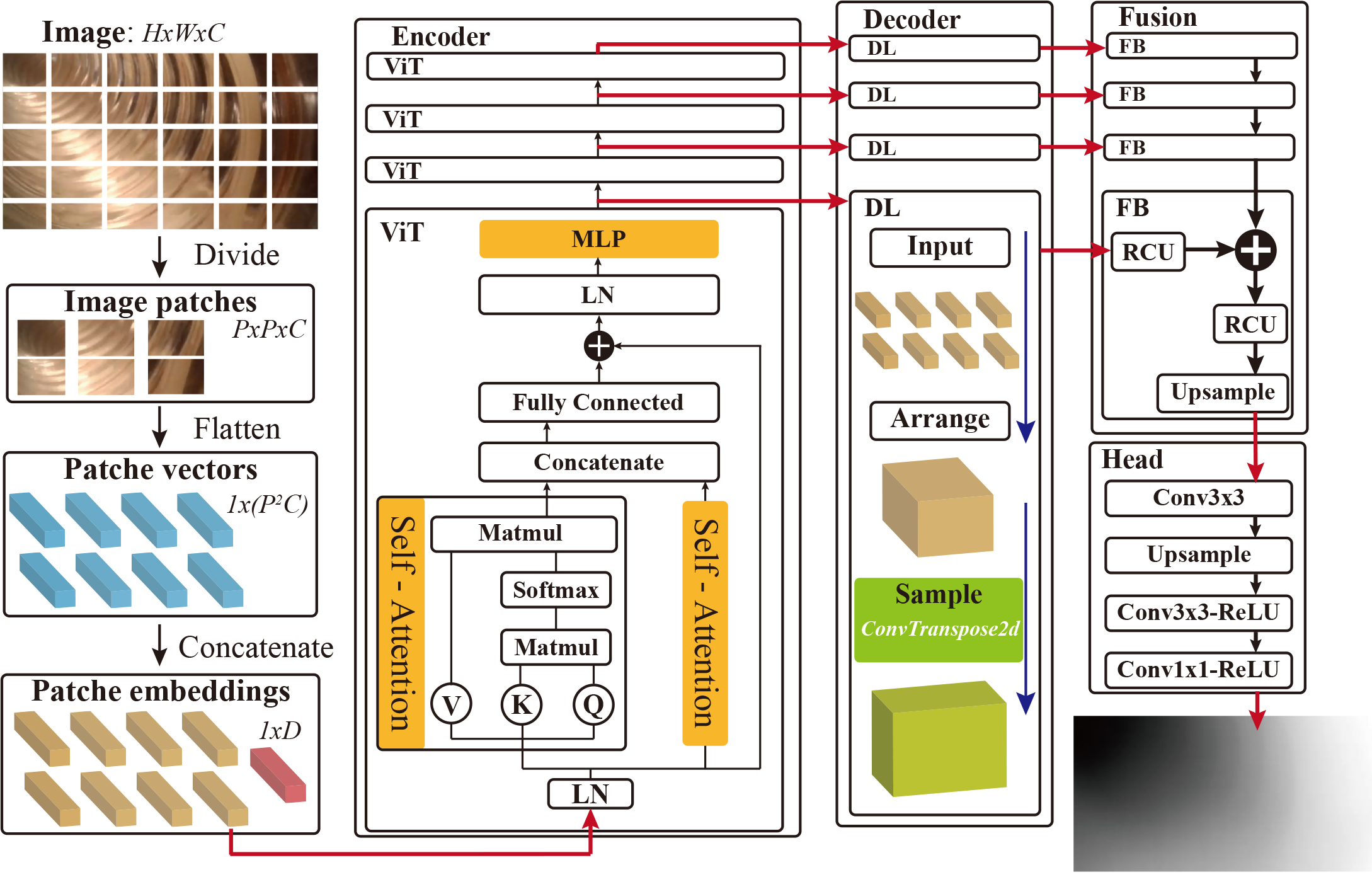}
  \setlength{\abovecaptionskip}{-0.4cm}
  \caption{The proposed ViT-based depth estimation for monocular visual guidance and feedback with endoscopic image as input and depth map as output.}
  \label{fig:transformer}
  \vspace{-0.5cm}
\end{figure}

Our ViT-based depth estimation network (VDEN) is designed on an encoder-decoder structure using ViT as the encoder basic block as shown in Fig. \ref{fig:transformer}. 
We rearrange the embeddings from the encoder into image-like feature maps and combine them from a three-stage decoder into the final dense depth.
The standard transformer receives as input a 1-D sequence of token embeddings, so we first map the endoscopic image $\bm{\mathcal{X}} \in \mathbb{R}^{H \times W \times C}$ into a stacked sequence $\bm{\mathcal{K}}_0 \in \mathbb{R}^{(N_p + 1) \times D}$ in matrix form, where $(H,W)$ denotes the resolution of the image, and $C$ is the number of channels. 
Specifically, the image $\bm{\mathcal{X}}$ is divided into $N_P$ patches with the size of ${P \times P \times C}$, where $(P,P)$ represents the resolution of each patch and $N_P = H \cdot W / P^2$. 
Then, these patches are flattened into vectors $\bm{x}_1, \bm{x}_2, ..., \bm{x}_{n} \in \mathbb{R}^{1 \times (P^2 \cdot C)}$, $n \in \{1, 2, ..., N_P\}$.
To obtain the spatial position of each patch, 
we concatenate these patches with a position embedding and a special projection embedding, and refer to the output of this concatenation as the patch embeddings. 
Thus, these patch embeddings can be calculated from $\bm{\mathcal{K}}_0$ as: 
\begin{equation}
\setlength{\abovedisplayskip}{4pt}
\setlength{\belowdisplayskip}{4pt}
    \begin{small}
    \bm{\mathcal{K}}_0 = 
    \begin{bmatrix}
        \bm{x}_0^{\intercal} 
        & \bm{E}^{^{\intercal}} \bm{x}_1^{\intercal} 
        & \bm{E}^{^{\intercal}} \bm{x}_2^{\intercal} 
        & \cdots 
        & \bm{E}^{^{\intercal}} \bm{x}_{N_P}^{\intercal}
    \end{bmatrix}^{\intercal}
    + \bm{E}_{pos}
    \end{small}
\end{equation}
where $\bm{x}_0$ is a learned embedding to aggregate features into the image level, 
$\bm{E} \in \mathbb{R}^{(P^2 \cdot C) \times D}$ is the projection embedding, and $\bm{E}_{pos} \in \mathbb{R}^{(N_p + 1) \times D}$ is the position embedding with $D$ being the feature dimension of each patch embedding.

The patch embeddings are input into ViT-based encoder with $L$ ViT layers, and they are converted to new feature maps $\bm{\mathcal{K}}_l$, $l\in \{1, ..., L\}$, which are the output of the $l$-th ViT layer. 
Here, we adopt the transformer encoder designed by \cite{dosovitskiy2020image}. 
Each encoder layer has multi-head self-attention and multi-layer perception. 
Therefore, the global image features can be efficiently extracted from the encoder. 
Afterwards, we build the decoder \cite{ranftl2021vision}, each layer of which for depth estimation consists of three parts:
\begin{equation}
\setlength{\abovedisplayskip}{4pt}
\setlength{\belowdisplayskip}{4pt}
\begin{small}
\begin{aligned}
    & \textbf{Input}: \mathbb{R}^{(N_p + 1) \times D} \rightarrow \mathbb{R}^{N_p \times D} \\
    & \textbf{Arrange}: \mathbb{R}^{N_p \times D} \rightarrow \mathbb{R}^{\frac{H}{P} \times \frac{W}{P} \times D} \\
    & \textbf{Sample}: \mathbb{R}^{\frac{H}{P} \times \frac{W}{P} \times D} \rightarrow \mathbb{R}^{\frac{H}{r} \times \frac{W}{r} \times D^{'}}
\end{aligned}
\end{small}
\end{equation}
where $r$ is the output size ratio of the feature map w.r.t. the input image and $D^{'}$ is the dimension of the decoder output.

After the decoder, we combine the extracted feature maps using a residual convolution unit-based fusion block and gradually upsample the map in each fusion stage.
Finally, a depth estimation output head is used to produce the depth maps.
In addition, we adopt the scale-invariant trimmed loss \cite{ranftl2020towards} to train the whole model.
The region of interest (ROI) $\bm{\Omega}$ is determined from the largest connected component of the deepest 5 \% pixels, as the navigation guidance.

\subsection{Eye-in-Hand Vision-Motor Configuration} 
Given the current ROI $\bm{\Omega}$ with its center of mass $\bm{p} \in \mathbb{R}^{2}$ from VDEN represented by the blue contour and green point in the monocular visual guidance module of Fig. \ref{fig:diagram}, respectively, a learning-based adaptive controller for online update of the flexible endoscope model together with eye-in-hand visual model, and simultaneous visual servoing, is designed to drive the image central region towards the current ROI $\bm{\Omega}$.
Therefore, the mass center $\bm{p}_d \in \mathbb{R}^{2}$ of the desired region $\bm{\Omega}_d$ (red contour) is located at the image center, which can be also chosen manually by the clinician.
we implement a region-based visual servoing method here, the details of which can be found in a previous work \cite{yang2019adaptive}. 
We accordingly define the composed image feature as $\bm{y} \in \mathbb{R}^{2}$, which represents the smoothed mass center coordinates of $\bm{\Omega}$ to ensure its continuity through winding number calculation.

To derive the visual servoing controller, we first define the vision-motor model with the task-space velocity, i.e., eye-in-hand camera velocity $\bm{v}_c \in \mathbb{R}^{6}$, and the motor position input of the endoscope system as $\bm{q} = \begin{bmatrix} q_1 & q_2 & q_3 \end{bmatrix}^{\intercal} \in \mathbb{R}^{3}$, where $q_1$ and $q_2$ represent the motor positions corresponding to deflections of the distal continuum mechanism, and $q_3$ is the motor position for insertion motion.
The differential kinematics of the flexible endoscope that relates the camera velocity $\bm{v}_c$ to the motor velocity $\dot{\bm{q}}$, is given as $\bm{v}_c = \bm{\mathcal{J}}_r (\bm{q}) \dot{\bm{q}}$, 
where $\bm{\mathcal{J}}_r (\bm{q}) \in \mathbb{R}^{6 \times 3}$ denotes the Jacobian matrix of the endoscope system.
The eye-in-hand visual model of the flexible endoscope mapping the image feature flow $\dot{\bm{y}}$ to the camera motion $\bm{v}_c$ can be formulated as $\dot{\bm{y}} = \bm{\mathcal{L}}_y \bm{v}_c$, 
where $\bm{\mathcal{L}}_y \in \mathbb{R}^{2 \times 6}$ represents the vision interaction matrix for the image feature $\bm{y}$. 
From above two mappings, we can derive the vision-motor configuration of the endoscope system as
\begin{equation}
\setlength{\abovedisplayskip}{3pt}
\setlength{\belowdisplayskip}{3pt}
    \label{eq:visJ}
    \begin{small}
    \dot{\bm{y}} = \bm{\mathcal{J}}_c (\bm{q})
    \dot{\bm{q}} = 
    \bm{\mathcal{L}}_y \bm{\mathcal{J}}_r (\bm{q}) 
    \dot{\bm{q}}
    \end{small}
\end{equation}
where $\bm{\mathcal{J}}_c (\bm{q}) = \bm{\mathcal{L}}_y \bm{\mathcal{J}}_r (\bm{q})  \in \mathbb{R}^{2 \times 3}$ denotes the combined image Jacobian matrix of flexible endoscope mapping its motor velocity $\dot{\bm{q}}$ to the image feature velocity $\dot{\bm{y}}$ (i.e., image feature flow).
Note that the endoscope Jacobian $\bm{\mathcal{J}}_r (\bm{q})$ is assumed to be unknown due to its nonlinear structural properties, hence we approximate the image Jacobian $\bm{\mathcal{J}}_c (\bm{q})$ without any priori identification of it.

\subsection{Learning-Based Adaptive Visual Servoing} 
To this end, we employ two adaptive NNs $\bm{\mathcal{W}}_i \bm{\theta}_i (\bm{q}), i \in \{1, 2\}$ as shown in Fig. \ref{fig:diagram}, to learn the rows of image Jacobian $\bm{\mathcal{J}}_c (\bm{q})$, with motor position $\bm{q}$ as input given by
\begin{equation}
\setlength{\abovedisplayskip}{4pt}
\setlength{\belowdisplayskip}{4pt}
    \label{eq:visJnn}
    \begin{small}
    \bm{\mathcal{J}}_c (\bm{q}) = 
    \begin{bmatrix}
        \bm{\mathcal{W}}_1 \bm{\theta}_1 (\bm{q}) & \bm{\mathcal{W}}_2 \bm{\theta}_2 (\bm{q}) 
    \end{bmatrix}^{\intercal}
    \end{small}
\end{equation}
where $\bm{\mathcal{W}}_i \in \mathbb{R}^{3 \times \xi}, i \in \{1, 2\}$, represents the ideal weight matrix of the $i$-th NN with $\xi$ neurons, and $\bm{\theta}_i (\bm{q}) \in \mathbb{R}^{\xi}, i \in \{1, 2\}$, denotes the corresponding vector of the activation functions.
With online learning of the image Jacobian in (\ref{eq:visJnn}), we can linearly parameterize the image feature flow $\dot{\bm{y}}$ as
\begin{equation}
\setlength{\abovedisplayskip}{4pt}
\setlength{\belowdisplayskip}{4pt}
    \label{eq:visFlow}
    \begin{small}
	\dot{\bm{y}} = 
	\underbrace{
	\begin{bmatrix}
        \bm{\mathcal{W}}_1 \bm{\theta}_1 (\bm{q})
        & \bm{\mathcal{W}}_2 \bm{\theta}_2 (\bm{q})
    \end{bmatrix}^{\intercal}
    }_{\bm{\mathcal{J}}_c (\bm{q})}
	\dot{\bm{q}}
	=
	\bm{\Theta}^{\intercal}(\bm{q}) 
	\bm{\mathcal{Q}}^{\intercal}(\dot{\bm{q}}) 
	\overline{\bm{\mathcal{W}}}
	\end{small}
\end{equation}
where $\bm{\Theta}(\bm{q}) = \mathrm{diag} ( \bm{\theta}_{1} (\bm{q}), \ \bm{\theta}_{2} (\bm{q}) ) \in \mathbb{R}^{2 \xi \times 2}$ 
and $\bm{Q} (\dot{\bm{q}}) = \mathrm{diag} ( \dot{\bm{q}}, \ \dot{\bm{q}}, \ \cdots, \ \dot{\bm{q}} ) \in \mathbb{R}^{6 \xi \times 2 \xi}$ are diagonal block matrices grouping the activation functions $\bm{\theta}_i (\bm{q})$, $i \in \{1, 2\}$, and $2 \xi$ motor motion $\dot{\bm{q}}$, respectively, and $\overline{\bm{\mathcal{W}}} \in \mathbb{R}^{6 \xi}$ denotes a vector stacking the columns of the ideal weights $\bm{\mathcal{W}}_i$, $i \in \{1, 2\}$.

Since the real image Jacobian $\bm{\mathcal{J}}_c (\bm{q})$ is unknown, we estimate it denoted by $\widehat{\bm{\mathcal{J}}}_c (\bm{q}) \in \mathbb{R}^{2 \times 3}$ through two estimated adaptive NNs $\widehat{\bm{\mathcal{W}}}_i \bm{\theta}_i (\bm{q}), i \in \{1, 2\}$, given by
\begin{equation}
\setlength{\abovedisplayskip}{4pt}
\setlength{\belowdisplayskip}{4pt}
    \label{eq:esti_visJnn}
    \begin{small}
    \widehat{\bm{\mathcal{J}}}_c (\bm{q}) =
    \begin{bmatrix}
        \widehat{\bm{\mathcal{W}}}_1 \bm{\theta}_1 (\bm{q})
        & \widehat{\bm{\mathcal{W}}}_2 \bm{\theta}_2 (\bm{q})
    \end{bmatrix}^{\intercal}
    \end{small}
\end{equation}
where $\widehat{\bm{\mathcal{W}}}_i \in \mathbb{R}^{3 \times \xi}, i \in \{1, 2\}$, represents the estimated NN weight matrix.
Therefore, we can derive the prediction error of image feature flow $\dot{\widetilde{\bm{y}}} \in \mathbb{R}^{2}$, i.e., the image flow error as
\begin{equation}
\setlength{\abovedisplayskip}{4pt}
\setlength{\belowdisplayskip}{4pt}
    \label{eq:esti_visFlowE}
    \begin{small}
    \dot{\widetilde{\bm{y}}}
    =  \dot{\bm{y}}
    - \widehat{\bm{\mathcal{J}}}_c (\bm{q}) \dot{\bm{q}}
    = \bm{\Theta}^{\intercal}(\bm{q}) \bm{\mathcal{Q}}^{\intercal}(\dot{\bm{q}})
    \widetilde{\overline{\bm{\mathcal{W}}}}
    \end{small}
\end{equation}
where $\widetilde{\overline{\bm{\mathcal{W}}}} = \overline{\bm{\mathcal{W}}} - \widehat{\overline{\bm{\mathcal{W}}}} \in \mathbb{R}^{6 \xi}$ denotes the NN weights estimation error in vector form, which can be updated from a modified composite adaptation algorithm \cite{slotine1991applied} as 
\begin{equation}
\setlength{\abovedisplayskip}{4pt}
\setlength{\belowdisplayskip}{4pt}
    \label{eq:adaptive}
    \begin{small}
    \dot{\widehat{\overline{\bm{\mathcal{W}}}}}
	=
	\bm{\Gamma}^{-1}_W
	\left[
	\mu_e \bm{\mathcal{Q}} (\dot{\bm{q}}) \bm{\Theta} (\bm{q}) 
	\bm{e} 
	+
	\mu_y \bm{\mathcal{Q}} (\dot{\bm{q}}) \bm{\Theta} (\bm{q}) 
	\dot{\widetilde{\bm{y}}}
    \right]
    \end{small}
\end{equation}
where $\mu_e$ and $\mu_y$ are positive constants, 
$\bm{\Gamma}_W \in \mathbb{R}^{6 \xi \times 6 \xi}$ is a positive definite and diagonal gain matrix, 
and $\bm{e} = \bm{y} - \bm{y}_d \in \mathbb{R}^{2}$ is denoted as the image feature error (tracking error). 

Finally, we can design a velocity controller to drive the image feature $\bm{y}$ towards the desired one $\bm{y}_d$ by using the learned image Jacobian $\widehat{\bm{\mathcal{J}}}_c (\bm{q})$ as $\dot{\bm{q}} = - \mu_c \widehat{\bm{\mathcal{J}}}^{+}_c (\bm{q}) \bm{e}$,
where $\mu_c$ is a positive constant and $\widehat{\bm{\mathcal{J}}}^{+}_c (\bm{q}) \in \mathbb{R}^{3 \times 2}$ denotes the Moore–Penrose pseudo-inverse of $\widehat{\bm{\mathcal{J}}}_c (\bm{q})$.
Under this control algorithm, the closed-loop system stability, together with the convergences of image feature error and image flow error to zeros, i.e., $\lim\limits_{t \to \infty} \bm{e} = 0$ and $ \lim\limits_{t \to \infty} \dot{\widetilde{\bm{y}}} = 0$, can be guaranteed, and a similar proof of which can be found in \cite{lu2022robust} for shape servoing tasks.
Alternatively, we can apply a learning-based MPC strategy to predict the future behavior of the flexible endoscope over a future time horizon $\Phi$ as
\begin{equation}
\setlength{\abovedisplayskip}{2pt}
\setlength{\belowdisplayskip}{4pt}
    \label{eq:visOpti}
    \begin{small}
    \argmin_{\dot{\bm{q}}(t\!+\!k\!+\!1)} \;
    \sum_{k = 0}^{\Phi} \eta_k
    \left\| \bm{y} (t\!+\!k\!+\!1) - \bm{y}_d \right\|^{2}_{2}
    \end{small}
\end{equation}
s.t. $\dot{\bm{q}}_3 (t\!+\!k) \ge 0$ as insertion motion with the predicted feature $\bm{y} (t\!+\!k\!+\!1)$ for $k \in \{0, 1, 2,..., \Phi \}$ being
\begin{equation}
\setlength{\abovedisplayskip}{5pt}
\setlength{\belowdisplayskip}{5pt}
    \label{eq:visCons1}
    \begin{small}
    \bm{y} (t\!+\!k\!+\!1) = \bm{y} (t\!+\!k) 
    + \widehat{\bm{\mathcal{J}}}_c (\bm{q} (t\!+\!k)) \dot{\bm{q}} (t\!+\!k) \Delta t   
    \end{small}
\end{equation}
where $\Delta t$ is the iteration interval, and the image Jacobian $\widehat{\bm{\mathcal{J}}}_c (\bm{q} (t\!+\!k))$, $k \in \{0, 1, 2,..., \Phi \}$, can be predicted through the adaptive NNs as shown in Fig. \ref{fig:diagram}.




\section{Energy-Motivated 3-D Shape Planning and Autonomous Navigation Framework}
In this section, we describe an energy-motivated shape planning algorithm by using FBG sensors, and a learning-based MPC framework for flexible endoscope navigation.

\subsection{FBG-based Proprioception of Elastic Potential Energy}
\begin{figure}[!t]
  \centering
  \includegraphics[width=\linewidth]{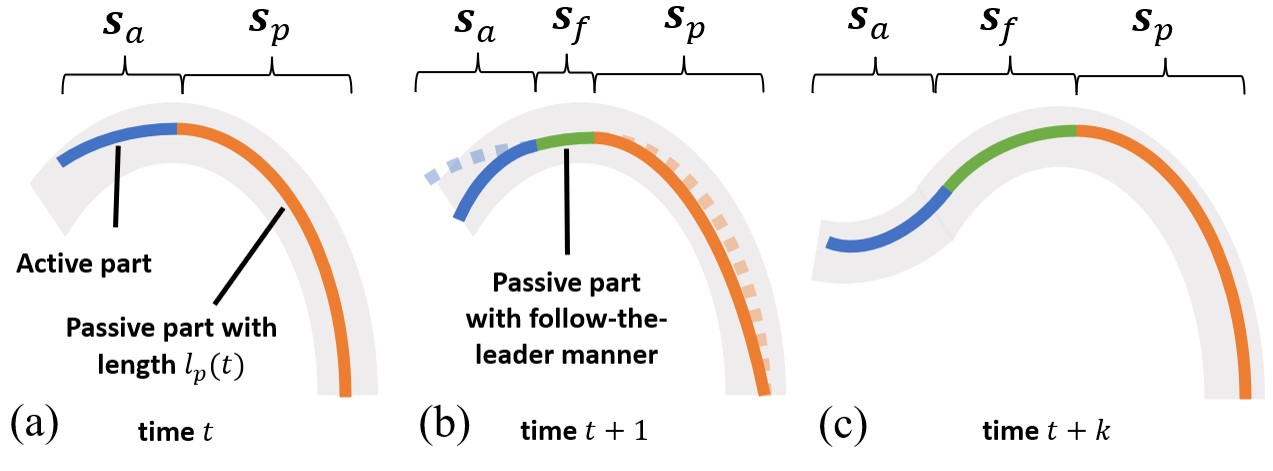}
  \setlength{\abovecaptionskip}{-0.5cm}
  \caption{Shape prediction of flexible endoscope for elastic potential energy minimization: (a) current shape, (b) shape at time $t\!+\!1$, and (c) shape at time $t\!+\!k$.}
  \label{fig:planning}
  \vspace{-0.6cm}
\end{figure}

To avoid unnecessary and excessive interactions with surrounding anatomy during endoscopic navigation, we design an optimization-based planning method for minimizing the elastic potential energy flow of the system.
By solely embedding a multi-core FBG fiber in the flexible endoscope, a robust and accurate filtering-based algorithm was previously reported \cite{lu2021robust} to acquire the endoscope 3-D shape, which is self-contained, independent of external sensors, and can maintain high sensing quality against perturbations. 
At current time $t$, we construct the entire endoscope shape by two part including the active part $\bm{s}_a \in \mathbb{R}^{m_a}$ of the distal continuum mechanism, and passive part $\bm{s}_p \in \mathbb{R}^{m_p}$ with a length of $l_p (t)$ as shown in Fig. \ref{fig:planning} (a).
In addition to shape sensing as shown in Fig. \ref{fig:diagram}, we utilized the bending and torsion signals from the distributed FBG sensors to approximate the elastic potential energy of the flexible endoscope formulated by $\mathcal{E} (t) = \mathcal{E} (\bm{s}_a (t), \; \bm{s}_p (t)): \mathbb{R}^{m_a} \times \mathbb{R}^{m_p} \mapsto \mathbb{R}$, following the working principal of strain measurements.

\begin{figure*}[!t]
  \centering
  \includegraphics[width=\linewidth]{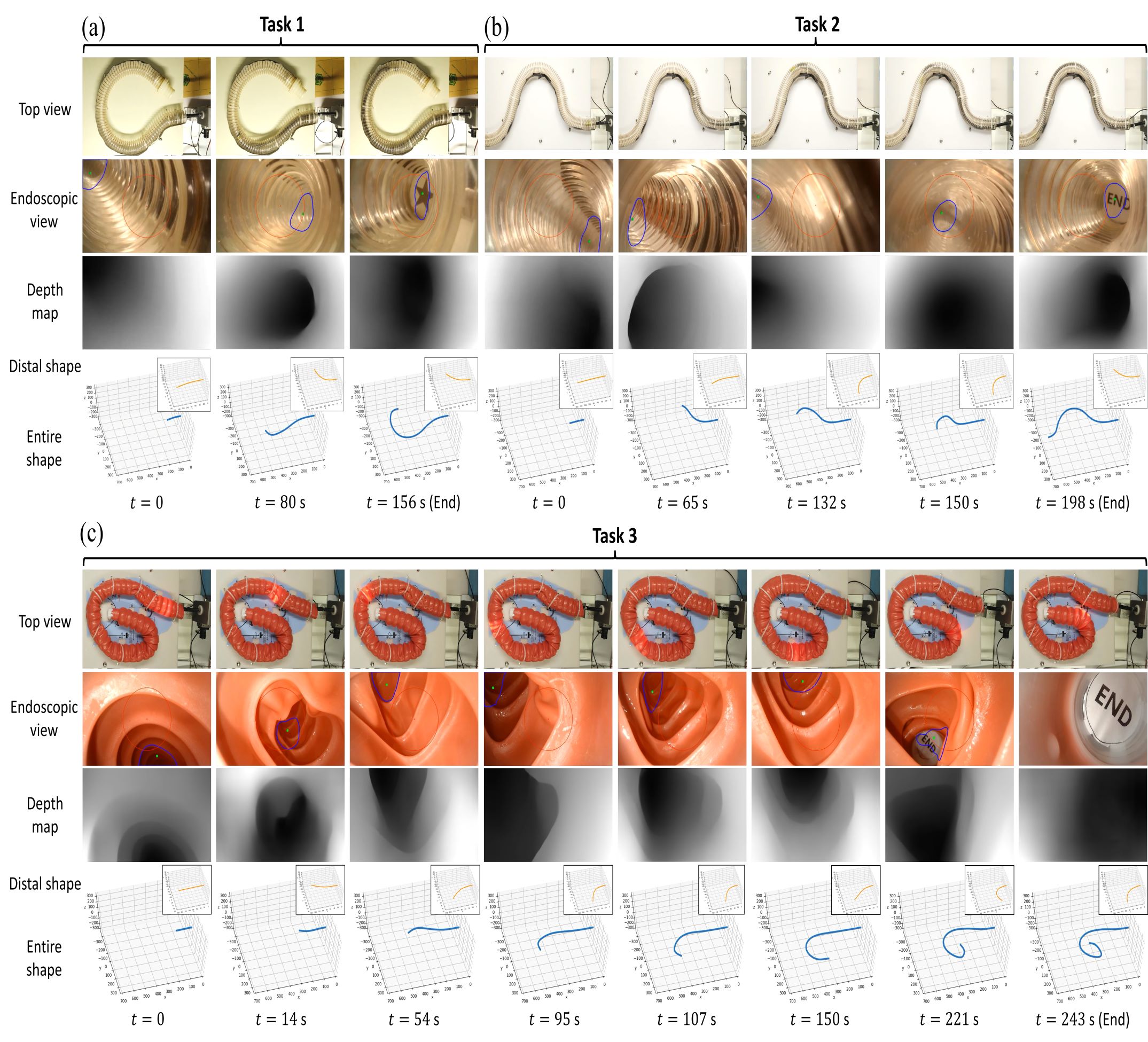}
  \setlength{\abovecaptionskip}{-0.5cm}
  \caption{Snapshots in Task 1, 2, and 3: top view, endoscopic view, depth map, and FBG shapes of entire colonoscope and distal part.}
  \label{fig:exp0}
  \vspace{-0.4cm}
\end{figure*}

To minimize the future potential energy flow, we predict the elastic potential energy $\mathcal{E} (t\!+\!k\!+\!1)$, $k \in \{0, 1, 2,..., \Phi \}$, by dividing the entire endoscope shape into three parts: the active part $\bm{s}_a (t\!+\!k\!+\!1)$, passive part with follow-the-leader manner $\bm{s}_f (t\!+\!k\!+\!1) \in \mathbb{R}^{m_f}$, where $m_f (t\!+\!k\!+\!1)$ is dependent on the insertion length at each time instant, and passive part $\bm{s}_p (t\!+\!k\!+\!1)$ with the length of $l_p(t)$, as shown in Fig. \ref{fig:planning} (b).
By taking advantage of state prediction in the Kalman filter, the sensing algorithm is hereby improved to predict the passive part shape $\bm{s}_p (t\!+\!k\!+\!1)$ as shown in the FBG-based sensing module of Fig. \ref{fig:diagram}, which is deformed due to its interactions with anatomical tissues during endoscope insertion. 
Using FBG-based proprioception of the endoscope 3-D configuration, the active part shape $\bm{s}_a (t\!+\!k\!+\!1)$ can be predicted through a learning-based approach as
\begin{equation}
\setlength{\abovedisplayskip}{4pt}
\setlength{\belowdisplayskip}{4pt}
    \label{eq:fbgCons1}
    \begin{small}
    \bm{s}_a (t\!+\!k\!+\!1) = \bm{s}_a (t\!+\!k)
    + \widehat{\bm{\mathcal{J}}}_s (\bm{q}^{'} (t\!+\!k)) \dot{\bm{q}}^{'} (t\!+\!k) \Delta t
    \end{small}
\end{equation}
where $\bm{q}^{'} = \begin{bmatrix} q_1 & q_2 \end{bmatrix}^{\intercal} \in \mathbb{R}^{2}$ denotes two motors' positions for distal deflections, $\widehat{\bm{\mathcal{J}}}_s (\bm{q^{'}}) \in \mathbb{R}^{m_a \times 2}$ is the estimated shape Jacobian corresponding to the actual one $\bm{\mathcal{J}}_s (\bm{q^{'}}) \in \mathbb{R}^{m_a \times 2}$, which relates the active part shape flow $\dot{\bm{s}}_a $ to $\dot{\bm{q}}^{'}$.
In a similar way updating the image Jacobian $\widehat{\bm{\mathcal{J}}}_c (\bm{q} (t\!+\!k))$, we can learn and predict the shape Jacobian $\widehat{\bm{\mathcal{J}}}_s (\bm{q}^{'} (t\!+\!k))$, from $m_a$ adaptive NNs \cite{lu2022robust} as shown in Fig. \ref{fig:diagram}.

\subsection{Potential Energy Prediction and Shape Planning}
By making the assumption of follow-the-leader behavior \cite{girerd2020slam,culmone2021follow}, we can predict the passive part shape $\bm{s}_f (t\!+\!k\!+\!1)$, $k \in \{0, 1, 2,..., \Phi \}$, following the current shape $\bm{s}_a (t)$ and the predicted ones $\bm{s}_a (t\!+\!k\!+\!1)$ of the endoscope active part as shown in Fig. \ref{fig:planning} (b), given by
\begin{equation}
\setlength{\abovedisplayskip}{4pt}
\setlength{\belowdisplayskip}{4pt}
    \label{eq:fbgCons2}
    \begin{footnotesize}
    \bm{s}_f (t\!+\!k\!+\!1) = 
    \begin{bmatrix}
        \bm{0}_{m_i \times m_o} 
        & \bm{I}_{m_i} 
        & \bm{0}_{m_i \times m_f} \\
        \bm{0}_{m_f \times m_o} 
        & \bm{0}_{m_f \times m_i} 
        & \bm{I}_{m_f} 
    \end{bmatrix}
    \begin{bmatrix}
        \bm{s}_a (t\!+\!k) \\
        \bm{s}_f (t\!+\!k)
    \end{bmatrix}
    \end{footnotesize}
\end{equation}
where $\bm{I}_{a} \in \mathbb{R}^{a \times a}$ and $\bm{0}_{a \times b} \in \mathbb{R}^{a \times b}$, $a \in \{m_i, m_f \}$ and $b \in \{m_i, m_f, m_o \}$, denote the identity and zero matrices, respectively, and $m_o = m_a - m_i$ with $m_i$ being the endoscope insertion length.
Consequently, the elastic potential energy of the flexible endoscope $\mathcal{E} (t\!+\!k^{'})$, $k^{'} \in \{1, 2,..., \Phi \}$, can be predicted from the entire endoscope shape (three parts) predicted at time $t\!+\!k^{'}$ as shown in Fig. \ref{fig:planning} (c), given by
\begin{equation}
\setlength{\abovedisplayskip}{3pt}
\setlength{\belowdisplayskip}{2pt}
\begin{small}
\begin{aligned}
    \label{eq:fbgCons3}
    \mathcal{E} (t\!+\!k^{'}) = \mathcal{E} \left( \bm{s}_a (t\!+\!k^{'}), \bm{s}_f (t\!+\!k^{'}), \bm{s}_p (t\!+\!k^{'}) \right) \\
    : \mathbb{R}^{m_a} \times \mathbb{R}^{m_f}\times \mathbb{R}^{m_p} \mapsto \mathbb{R}
\end{aligned}
\end{small}
\end{equation}
Therefore, a local shape trajectory with the smallest cost of the potential energy flow can be obtained by solving the following optimization problem as
\begin{equation}
\setlength{\abovedisplayskip}{2pt}
\setlength{\belowdisplayskip}{4pt}
    \begin{small}
    \argmin_{\dot{\bm{q}}(t\!+\!k\!+\!1)} \;
    \sum_{k = 0}^{\Phi} \lambda_k
    \left\| \mathcal{E} (t\!+\!k\!+\!1) - \mathcal{E} (t\!+\!k) \right\|^{2}_{2}
    \end{small}
\end{equation}
s.t. (\ref{eq:fbgCons1}), (\ref{eq:fbgCons2}), (\ref{eq:fbgCons3}),  and $\dot{\bm{q}}_3 (t\!+\!k) \ge 0$.
Note that the proposed shape planning algorithm is generic and feasible for other shape sensing techniques \cite{shi2016shape} rather than only for FBGs.



\subsection{Learning-Based Model Predictive Control}
In a combination of the depth-guided visual control and energy-motivated shape planning described above, we derive a data-driven framework for autonomous navigation of flexible endoscopes as shown in Fig. \ref{fig:diagram}, based on the development of a learning-based MPC strategy with the following formulation:
\begin{equation}
\setlength{\abovedisplayskip}{4pt}
\setlength{\belowdisplayskip}{4pt}
\begin{small}
\begin{aligned}
    \argmin_{\dot{\bm{q}}(t\!+\!k\!+\!1)} \;
    & \sum_{k = 0}^{\Phi} 
    \Big \{ \eta_k
    \left\| \bm{y} (t\!+\!k\!+\!1) - \bm{y}_d \right\|^{2}_{2}  \\
    & \quad + \lambda_k
    \left\| \mathcal{E} (t\!+\!k\!+\!1) - \mathcal{E} (t\!+\!k) \right\|^{2}_{2} \Big \}
\end{aligned}
\end{small}
\end{equation}
s.t. (\ref{eq:visCons1}) $\sim$ (\ref{eq:fbgCons3}), and $\dot{\bm{q}}_3 (t\!+\!k) \ge 0$, where the first term is to control the image ROI, the second term is for potential energy flow minimization, $\eta_k$ and $\lambda_k$, $k \in \{0, 1, 2,..., \Phi \}$, denote the weights w.r.t. these two tasks.
Finally, the steering policy over a time horizon $\dot{\bm{q}}(t\!+\!k\!+\!1)$, $k \in \{0, 1, 2,..., \Phi \}$, can be optimized by satisfying the above objective function s.t. the constraints, and then input $\dot{\bm{q}}(t\!+\!k\!+\!1)$ to the motors of the endoscope system, which can impose the desired image ROI and simultaneously minimize the potential energy flow.

\section{Experiments}
In this section, the experimental setup is firstly described, and the results are accordingly presented and discussed.

\subsection{Setup}

The training dataset for the proposed depth estimation is formulated with MIX 5 \cite{ranftl2020towards} and endoscopic images \cite{ozyoruk2021endoslam}.
The endoscopic images were constructed from virtual capsule endoscopy, which was developed in Unity and had 21887 frames with corresponding ground truth depth maps.
Our VDEN model was implemented in PyTorch and trained using Adam solver for 60 epochs with the batch size of 6 and the learning rate was set as le-5.
In the experiments, the original image was cropped to 384 $\times$ 384 $\times$ 3 as input.
The resolution $P=16$, the number $L$ of ViT layers was 24, and the dimensions $D = 768$ and $D^{'} = 256$ with the ratio $r=2$.

We performed the experiments on a robotic-assisted flexible endoscope system previously detailed in \cite{lu2021robust}, which was integrated with a colonoscope, a handle drive module for distal deflections, a dual-wheel friction module for endoscope insertion, a multi-core FBG fiber, and an eye-in-hand camera.
The system has a distal active part of 120 mm length and a passive working length of 880 mm, hence the FBG-based proprioception unit has 1 m sensing length in total.
The parameters in the MPC framework were set as follows: $\mu_e = 0.01$, $\mu_y = 0.2$, $\bm{\Gamma}_W^{-1} = \bm{I}_{54} \in \mathbb{R}^{54 \times 54}$, $\eta_k = 1 / 2^{k}$, $\lambda_k = 1 / 2^{k+1} $, $k \in \{0, 1, 2,..., \Phi \}$, and $\Phi = 20$, which were tuned according to the initial experiments with convergences of control and prediction errors to zeros.
As same with those in \cite{lu2022robust}, 9 neurons (i.e., $\xi = 9$) and radial basis functions (RBFs) were employed for NNs learning of both vision and shape prediction, and so were the center and width values.

\begin{figure}[!t]
  \centering
  \includegraphics[width=\linewidth]{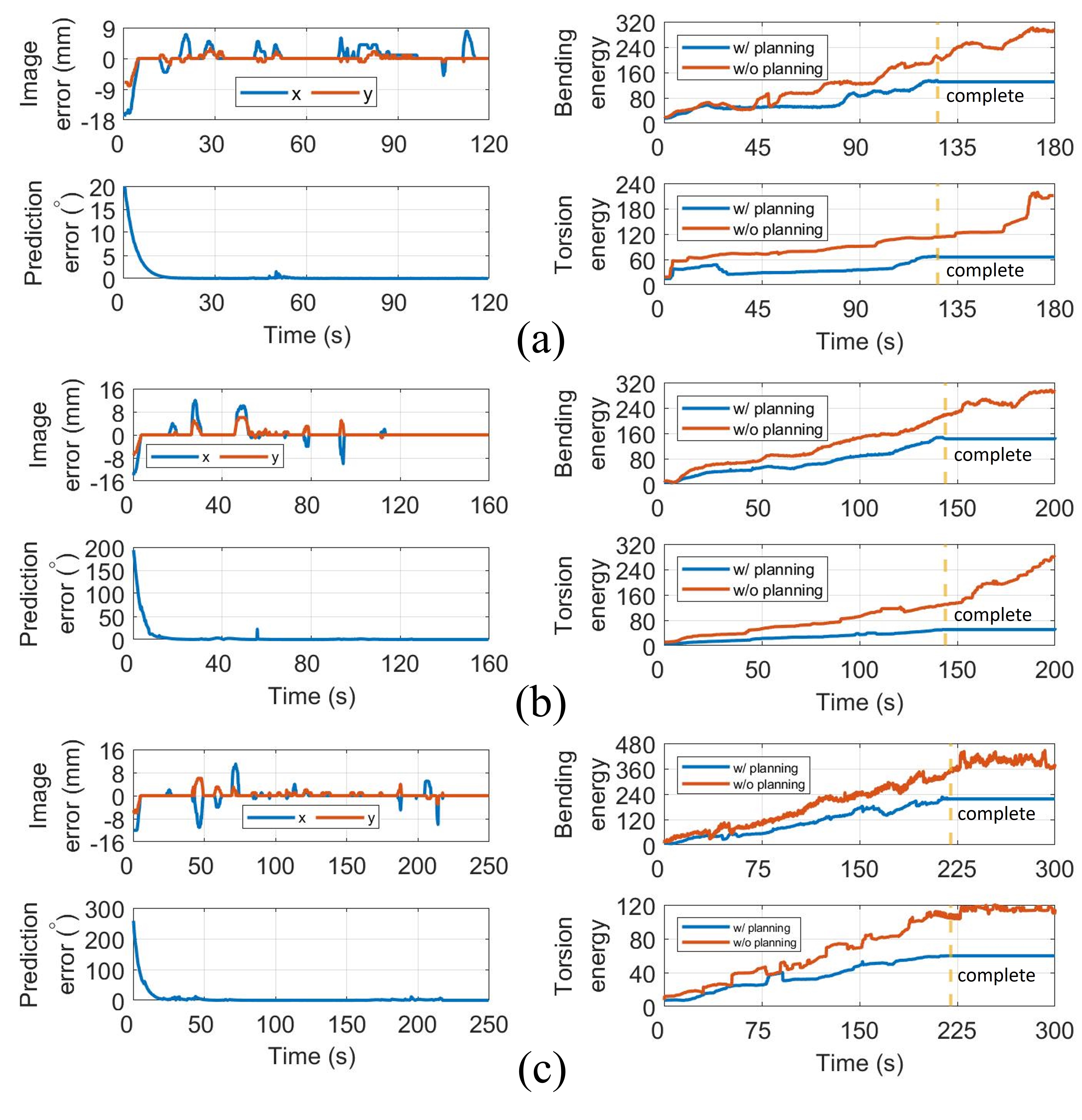}
  \setlength{\abovecaptionskip}{-0.5cm}
  \caption{Image tracking error, shape flow prediction error, bending and torsion potential energy using the methods with and without shape planning in (a) Task 1, (b) Task 2, and (c) Task 3.}
  \label{fig:exp3}
  \vspace{-0.4cm}
\end{figure}


\subsection{Results and Discussion}
To evaluate the performance of the proposed framework, three tasks compared with the method using vision guidance but without shape planning, were performed in two feature-less phantoms with different shapes and one colonoscopy phantom (Kyoto Kagaku M40) as shown in Fig. \ref{fig:exp0}. 
The phantoms have diameters of 50 mm and lengths of more than 1 m to mimic the endoscopic scenarios.
We conducted each task 8 times and recorded the results with snapshots in Fig. \ref{fig:exp0}, including the top view of endoscope, endoscopic view, depth map, and FBG-based shapes of entire colonoscope and distal part.
Four qualitative results during one trial of each task are plotted in Fig. \ref{fig:exp3} (a), (b), and (c), respectively, including image tracking error $\bm{e}$, 2-norm prediction error of shape flow $\| \dot{\widetilde{\bm{s}}}_a \|$, bending potential energy $\mathcal{E}_b$, and torsion potential energy $\mathcal{E}_t$.
Moreover, we present three quantitative metrics referring to \cite{pore2022colonoscopy}, in terms of insertion time $T_{in}$, endpoint trajectory length $L_{et}$, and average 2-norm of image tracking error $\left\| \bm{e} \right\|$ of three tasks in Table \ref{table:errors3}. 

For the performance of depth estimation, our VDEN model running at 19 FPS, can provide reasonable depth maps with accuracy of 90.88 \% in Task 1, 91.08 \% in Task 2, and 82.25 \% in Task 3, where the overlap degree of the estimated and real deep regions in each frame was examined.
Moreover, sharp boundaries were well kept, and the estimated depth increased smoothly from near to far scenes as demonstrated in the depth maps of Fig. \ref{fig:exp0}, despite in unknown environments with reflective and translucent surfaces.
We can observe the estimated depth maps during all tasks were particularly stable, hence helpful for autonomous endoscope navigation.

\begin{table}[!t]
\setlength{\abovecaptionskip}{0cm}
\setlength{\belowcaptionskip}{-0cm}
\captionsetup{font={small}}
\caption{\centering Comparison results of Task 1, 2, and 3}
\scriptsize
\label{table:errors3}
\centering
\begin{tabular}{c|c|ccc}
\hline
                        & Methods      & $T_{in}$ (s) & $L_{et}$ (mm)      & $\left\| \bm{e} \right\|$ (mm)    \\ \hline \hline
\multirow{2}{*}{Task 1} & w/o planning & 179 $\pm$ 26 & 1752.4 $\pm$ 449.3 & 1.24 $\pm$ 0.27                  \\ \cline{2-5} 
                        & w/ planning  & 125 $\pm$ 11 & 1181.0 $\pm$ 153.1 & 0.70 $\pm$ 0.29                   \\ \hline \hline
\multirow{2}{*}{Task 2} & w/o planning & 204 $\pm$ 13 & 1668.3 $\pm$ 141.4 & 1.25 $\pm$ 0.25                  \\ \cline{2-5} 
                        & w/ planning  & 159 $\pm$ 37 & 1292.4 $\pm$ 173.2 & 0.63 $\pm$ 0.23                  \\ \hline \hline
\multirow{2}{*}{Task 3} & w/o planning & 334 $\pm$ 22 & 1935.5 $\pm$ 397.1 & 1.47 $\pm$ 0.21 \\ \cline{2-5} 
                        & w/ planning  & 238 $\pm$ 17 & 1485.6 $\pm$ 119.9 & 0.87 $\pm$ 0.27 \\ \hline 
\end{tabular}
\vspace{-0.4cm}
\end{table}

As indicated in the endoscopic views of Fig. \ref{fig:exp0} as well as the image tracking errors in the top-left images of Fig. \ref{fig:exp3}, the learning-based eye-in-hand visual servoing can robustly and adaptively track the image ROI (blue contour) with average image errors of 0.70, 0.63, and 0.87 mm in Task 1, 2, and 3, respectively, and complete the navigation tasks.
It is noticed that the shape flow prediction errors converged to zeros as shown in the bottom-left plots of Fig. \ref{fig:exp3} (a), (b), and (c), which indicate that the NN weights were learned from their mapping in \cite{lu2022robust}.
Besides, the bending and torsion potential energy in the right plots of Fig. \ref{fig:exp3} (a), (b), and (c), demonstrate that our planning algorithm can finish the navigation tasks using much less elastic potential energy as compared with that without shape planning, which can prevent unnecessary and excessive elastic contact forces on surrounding anatomy.
From Table \ref{table:errors3}, the proposed method costs less time $T_{in}$ with shorter endpoint trajectory $L_{et}$ for navigation compared with the method without shape planning, which also outperforms the manually-controlled steering procedure (more than 15 min).
These results of phantom experiments support our claims that the framework can perform robust compliance to deformable environments, online learn the time-varying model, and simultaneously steer the flexible endoscope for efficient and safe navigation.

\section{Conclusions}
In this paper, we propose a novel data-driven autonomous navigation framework for flexible endoscopy, by leveraging monocular depth guidance and energy-motivated shape planning.
The method can online learn the eye-in-hand vision-motor configuration of flexible endoscope without any priori knowledge of system model and environmental information, and also minimize the system potential energy flow.
The results of several phantom experiments show the feasibility and adaptability of the framework.

In the future, we will optimize the framework with advanced perception and control approaches to improve its performance, which will be evaluated in more realistic scenes.
The system issues about learning approximation errors and colliding with anatomical tissues will be taken into consideration as well.

\newpage

\addtolength{\textheight}{-12cm}   

\bibliographystyle{IEEEtran}
\bibliography{./references}

\end{document}